\documentclass[runningheads]{llncs}

\usepackage{eccv}

\usepackage{eccvabbrv}
\usepackage{svg}
\usepackage{graphicx}
\usepackage{booktabs}
\usepackage{makecell}
\usepackage[table]{xcolor}
\usepackage{microtype}
\usepackage{tabularx}
\usepackage{multirow} 
\usepackage{subcaption} 
\usepackage{dsfont}
\usepackage[accsupp]{axessibility}  

\usepackage{array}         

\usepackage[most]{tcolorbox}
\newtcolorbox{rqbox}{
  colback=blue!5,
  colframe=blue!60,
  boxrule=0pt,
  borderline west={2pt}{0pt}{blue!80}, 
  left=8pt, right=6pt, top=6pt, bottom=6pt
}
\newtcolorbox{takeawaybox}{
  colback=orange!8,
  colframe=orange!70!black,
  boxrule=0pt,
  borderline west={2pt}{0pt}{orange!85!black},
  left=8pt, right=6pt, top=6pt, bottom=6pt
}

\newcolumntype{L}[1]{>{\raggedright\arraybackslash}p{#1}}

\def\eg{\emph{e.g}\onedot} 
\def\ie{\emph{i.e}\onedot}

\definecolor{U_color}{RGB}{213, 26, 139}
\definecolor{M_color}{RGB}{251, 174, 66}
\definecolor{S1_color}{RGB}{64, 151, 245}
\definecolor{S2_color}{RGB}{69, 208, 55}

\usepackage{hyperref}

\usepackage{orcidlink}

\begin{document}


\title{\includegraphics[height=1.6em]{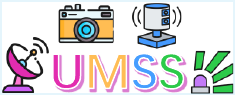} \hspace{0pt}: Towards \textcolor{U_color}{U}nsupervised \textcolor{M_color}{M}ulti-modal 
\textcolor{S1_color}{S}emantic \textcolor{S2_color}{S}egmentation
}

\titlerunning{Towards Unsupervised Multi-modal Semantic Segmentation}

\author{Haitian Zhang\inst{1}\orcidlink{0009-0005-9151-7543} \and
Thai Duy Nguyen\inst{1}\orcidlink{0009-0009-5287-9962} \and
Xiangyuan Wang\inst{2}\orcidlink{0009-0006-8446-2254} \and
Mohan Liu\inst{1}\orcidlink{0009-0003-2072-8509} \and
Lin Wang\inst{1}\orcidlink{0000-0002-7485-4493}\thanks{Corresponding author.}
}

\authorrunning{H.~Zhang et al.}

\institute{
EmPACT Lab, School of EEE, Nanyang Technological University, Singapore\\
\email{\{haitian003,nguyendu003\}@e.ntu.edu.sg}\\
\email{\{mohan.liu,linwang\}@ntu.edu.sg}
\and
The University of Hong Kong, Hong Kong SAR, China\\
\email{xiangyuan.wang@connect.hku.hk} \\
\url{https://empactlab.github.io/UMSS/}
}

\maketitle

\begin{abstract}

Multi-modal semantic segmentation (MSS) is essential for robust perception in complex environments, yet its potential remains largely untapped due to the prohibitive cost of human annotations. 
While unsupervised semantic segmentation (USS) has seen success on single RGB modality, \textbf{its naive extension to multi-modal data is hampered by fusion degradation}.
This is because, in the absence of explicit supervision, existing frameworks struggle to reconcile the heterogeneous structural patterns captured by different sensors, failing to effectively exploit their complementary information. In this paper, we make the \textbf{first} attempt to address the novel problem of \textbf{Unsupervised Multi-modal Semantic Segmentation (UMSS)}, aiming to effectively exploit complementary sensor information in a fully label-free setting. To this end, we propose \textbf{UniM2} (\textbf{Uni}fied \textbf{M}ulti-\textbf{M}odal), a novel framework built upon DINOv3 that transforms conventional fusion methods into consistent performance gains. \textbf{Our key idea} is to learn a unified latent space driven by \textbf{Cross-modal Correspondence Synergy (CMCS)} to extract intrinsic shared semantic cues, bypassing the need for label-guided adaptive fusion. To mitigate inherent inter-modal conflicts, we introduce a \textbf{Cross-modal Harmonizer (CMH)} that designates RGB as a stable reference, effectively suppressing inconsistent relational supervision while guiding the model to exploit complementary structural features. 
Extensive experimental results on NYU-Depth-v2 and MFNet show that \textbf{UniM2} improves mIoU by \textbf{6.4\%} and \textbf{9.8\%}, respectively, demonstrating clear advantages over existing frameworks in UMSS task.

\end{abstract}


\vspace{-10pt}
\section{Introduction}
\vspace{-10pt}

\begin{figure*}[t!]
    \centering
    \begin{subfigure}[b]{0.51\textwidth}
        \centering
        \includegraphics[width=\textwidth]{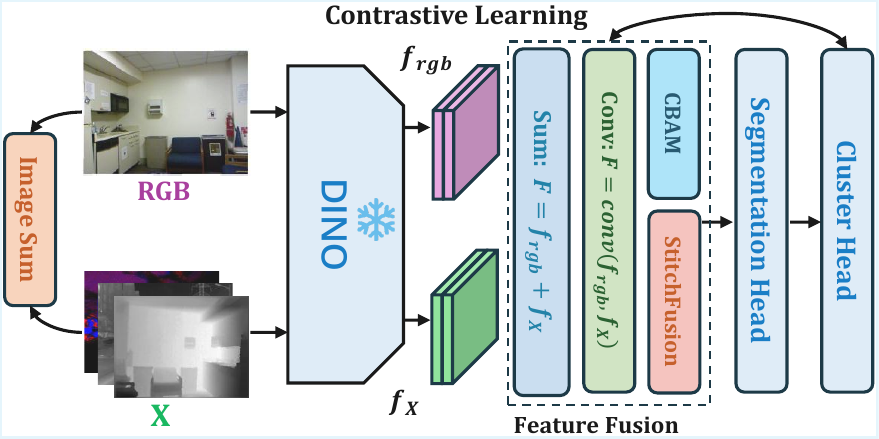}
        \caption{Directly Apply Conventional Fusion in UMSS}
        \label{fig:fusion_methods}
    \end{subfigure}
    \hfill
    \begin{subfigure}[b]{0.48\textwidth}
        \centering
        \includegraphics[width=\textwidth]{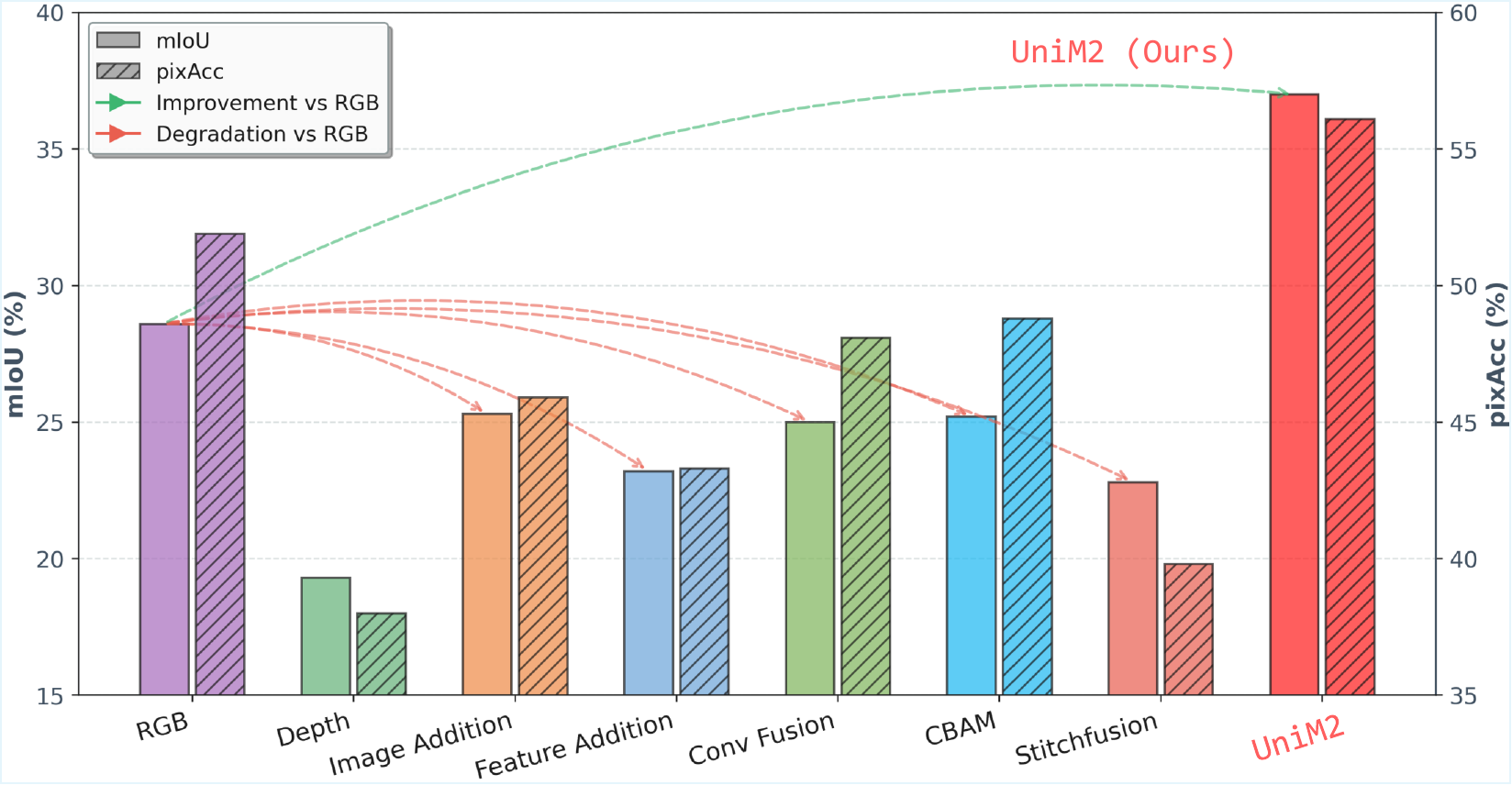}
        \caption{Performance Comparison}
        \label{fig:tissue}
    \end{subfigure}
    \vspace{-15pt}
    \caption{\textbf{Analysis of multi-modal integration in unsupervised semantic segmentation}. (a) We explore various fusion schemes, including naive \textbf{Image Addition}, \textbf{Feature Addition}, and \textbf{Conv Fusion}, as well as SOTA fusion methods in MSS such as \textbf{CBAM} \cite{CBAM} and \textbf{StitchFusion} \cite{Stitchfusion}. (b) Quantitative results on NYU-Depth-v2 \cite{NYU_Depth} demonstrate that existing advanced fusion strategies in multi-modal segmentation inevitably lead to performance degradation compared to the single RGB baseline in the unsupervised setting, while our \textbf{UniM2} achieves significant mIoU gains.}
    \vspace{-15pt}
    \label{fig:preliminary_study}
\end{figure*}

Multi-modal semantic segmentation (MSS) \cite{Stitchfusion,CMNEXT,Geminifusion,MAGIC} is crucial for robust perception in various safety-critical applications, including autonomous driving \cite{driving1,driving2,driving3}, robotic navigation \cite{navigation1,navigation2}, and embodied intelligence \cite{embodied1,embodied2,embodied3}. By integrating complementary signals \cite{NYU_Depth, DEOE, MFNET, Lidar_survey} such as depth \cite{NYU_Depth} or thermal/infrared \cite{FMB,MFNET}, MSS enhances perception in challenging environments where RGB-only perception often fail \cite{CV_survey1,CV_survey2}. Despite its importance, the progress of MSS is largely driven by massive human-annotated datasets \cite{COCO,deng2009imagenet}. These pixel-level labels are not only prohibitively expensive to produce but also constrain the learning process to a limited set of predefined semantic categories \cite{Unsupervised1,bojanowski2017unsupervised,sinaga2020unsupervised}, creating a barrier to utilizing the vast amounts of uncurated multi-modal data in label-free settings \cite{Unsupervised1,Unsupervised2,Unsupervised3}.

To bridge the gap between label-free learning and multi-modal perception, we define the task of \textbf{Unsupervised Multi-modal Semantic Segmentation (UMSS)}. A straightforward way to address this task is to extend SOTA multi-modal fusion strategies \cite{Stitchfusion,CMNEXT} to the USS framework \cite{STEGO, SmooSeg, HP, EAGLE, IL2Vseg} as illustrated in Fig.~\ref{fig:preliminary_study} (a). Although USS has achieved remarkable success on the single RGB modality propelled by the development of self-supervised Vision Transformers like the DINO family \cite{DINOV1,DINOV2,DINOV3}, simply extending these methods to multi-modal settings often results in performance degradation instead of the expected gains as shown in Fig.~\ref{fig:preliminary_study} (b). This phenomenon occurs because
\textit{existing frameworks struggle to reconcile the heterogeneous structural patterns and representation biases captured by different sensors without explicit guidance.}
While ground-truth annotations in supervised learning implicitly arbitrate these inter-modal inconsistencies, such \textbf{Conflicting Signals} cannot be properly resolved in an unsupervised setting. The lack of label-driven arbitration disturbs optimization and leads to a disorganized latent space with degraded clustering quality, which prevents the effective use of complementary information.

To address these challenges, we propose \textbf{UniM2}, a unified framework designed to learn a shared latent space driven by \textbf{Cross-modal Correspondence Synergy (CMCS)} as intrinsic supervision (Sec. \ref{sec:UniM2}). Instead of enforcing rigid feature alignment, CMCS promotes structural consistency across modalities by encouraging agreement in cross-modal correspondences, enabling the discovery of shared semantic manifolds while preserving complementary cues. To further mitigate inter-modal conflicts arising from heterogeneous sensing mechanisms, we designate RGB as the primary semantic reference and introduce a \textbf{Cross-modal Harmonizer (CMH)} (Sec. \ref{sec:CMH}). The CMH adaptively regulates alignment strength, suppressing unreliable relational supervision while retaining informative auxiliary signals, thereby preventing negative transfer during fusion.

We evaluate \textbf{UniM2} on three representative multi-modal benchmarks using the latest \textbf{DINOv3} \cite{DINOV3} architecture. Specifically, we investigate the ``R+X'' setting on bi-modal datasets, \ie, \textbf{NYU-Depth-v2} \cite{NYU_Depth} and \textbf{MFNet} \cite{MFNET}, and further validate the scalability of our framework on the quad-modal \textbf{MCubeS} \cite{McubeS} dataset. Experimental results show that conventional supervised fusion strategies fail to fully exploit the \textbf{Inherent Complementarity} of heterogeneous modalities in the absence of labels, resulting in severe performance degradation. In contrast, \textbf{UniM2} achieves absolute mIoU gains of \textbf{6.4\%} on NYU-Depth-v2 and \textbf{9.8\%} on MFNet over the RGB-only baseline, consistently transforming cross-modal interference into synergistic improvements. Furthermore, the modular design of CMH facilitates its extension to $N$ auxiliary modalities, yielding incremental gains on MCubeS. Our contributions are summarized as follows:
\begin{itemize}
    \item \textbf{Task Definition.} We introduce the task of Unsupervised Multi-modal Semantic Segmentation, investigating how to leverage heterogeneous modalities for semantic segmentation without any human annotations.
    \item \textbf{UniM2 Framework.} We propose UniM2, learning a unified latent space via Cross-modal Correspondence Synergy while utilizing a Cross-modal Harmonizer to regulate alignment strength and mitigate modal conflicts.
    \item \textbf{Performance and Scalability.} Built upon DINOv3, UniM2 converts the performance degradation typical of conventional fusion into substantial mIoU gains, while naturally scaling to multiple auxiliary modalities.
\end{itemize}
\label{sec:intro}

\vspace{-15pt}
\section{Related Work}
\vspace{-10pt}
\textbf{Multi-modal Semantic Segmentation.}
Multi-modal semantic segmentation \cite{CMNEXT,MAGIC,Stitchfusion} leverages complementary data from heterogeneous sensors to overcome the inherent limitations of single-modal RGB perception, particularly in visually degraded environments \cite{FAOD,zhou2025dfvo,shin2023deep}. Existing methods typically design modality-aware fusion mechanisms \cite{CMNEXT} to integrate heterogeneous features, including hierarchical feature aggregation, cross-modal attention \cite{CBAM}, and adaptive gating strategies \cite{MAGIC}. These approaches rely on dense pixel-level annotations to learn effective modality alignment, resolve cross-modal inconsistencies, and suppress conflicting predictions during training. While such supervised paradigms have demonstrated strong performance gains, their reliance on explicit semantic labels prevents direct extension to the UMSS settings, where no annotation is available to guide modality interaction. \textit{In fact, directly embedding supervised multi-modal fusion modules into existing USS frameworks not only fails to bring improvements, but often leads to substantial performance degradation.}

\noindent \textbf{Unsupervised Semantic Segmentation.}
Unsupervised semantic segmentation \cite{STEGO,EAGLE} aims to partition images into semantically meaningful regions without human-provided labels. Early studies \cite{DC,IIC,PiCIE} focused on discovering recurring visual patterns using hand-crafted features or low-level spatial priors such as pixel consistency and spatial continuity. However, these approaches were limited in capturing complex semantic variations due to insufficient high-level representation capacity. The development of self-supervised Vision Transformers (ViTs), particularly the DINO family \cite{DINOV1,DINOV2,DINOV3}, significantly advanced USS by providing semantically structured dense representations through large-scale pre-training. Building upon these representations, STEGO \cite{STEGO} introduced a distillation-based framework that converts dense feature correlations into discrete semantic maps via contrastive learning \cite{chuang2020debiased,khosla2020supervised}. Subsequent works \cite{SmooSeg,HP,EAGLE,IL2Vseg} further refined this paradigm by improving clustering objectives and exploiting local structural priors to enhance boundary quality. \textit{Despite these advances, existing USS methods are limited to RGB-only input and do not explicitly consider heterogeneous sensors}\cite{NYU_Depth,MUSE,McubeS,FMB,CMNEXT}. Consequently, the integration of complementary multi-modal signals in unsupervised settings remains largely unexplored.

\noindent \textbf{Representation Decoupling in Multi-modal Learning.}
Multi-modal learning \cite{multi-modal1,multi-modal3} aims to isolate modality-specific private information from shared commonalities to establish a robust semantic space \cite{decoup3,decoup4}. In supervised scenarios, this decoupling is inherently \textbf{label-driven} \cite{decoup1,decoup2}, as task-specific annotations explicitly guide the model to identify beneficial features while suppressing noise \cite{multi-modal2,multi-modal4}. \textit{Conversely, in unsupervised settings, decoupling becomes \textbf{highly unstable}; the lack of guidance often leads to optimization confusion, where structural contradictions between heterogeneous sensors degrade the shared manifold.} To address this instability, our \textbf{Cross-modal Harmonizer} provides \textbf{structured decoupling} by adaptively regulating alignment strength. Unlike naive fusion mechanisms, CMH suppress unreliable relational noise while concurrently harvesting complementary cues, ensuring that modality-specific nuances are leveraged without compromising the integrity of the latent semantic structure.

\noindent \textbf{Cross-modal/domain Adaptation via Distillation.}
Cross-modal knowledge distillation (CMKD) \cite{CMKD1,CMKD2,CMKD3} and Unsupervised Domain Adaptation (UDA) \cite{UDA1,UDA2,UDA3} share conceptual overlap with UMSS in leveraging multiple data sources. Typically, CMKD aims to transfer complementary information from an auxiliary modality to a primary one by guiding a student to imitate teacher signals \cite{distillation2}. This paradigm has been widely applied in cross-sensor perception such as event, LiDAR, and thermal transfer \cite{CMKD_event, CMKD_Lidar, CMKD_Depth, CMKD_infrared} to resolve spatial or illumination ambiguities. Similarly, UDA \cite{uda4,UDA5} focuses on bridging different data distributions through feature alignment or adversarial learning.

By contrast, UMSS fundamentally differs from both CMKD and UDA in two key aspects. First, the \textbf{supervision paradigm}. Unlike UDA which relies on labeled source domains or CMKD which depends on pre-trained ``teacher'' models \cite{distillation2}, UniM2 leverages \textbf{Cross-modal Correspondence Synergy} as an intrinsic, label-free supervisory signal. Second, the \textbf{learning objective}. While CMKD and UDA often focus on a ``teacher-student'' hierarchy \cite{distillation1} or source-to-target alignment to boost a primary modality, UniM2 treats heterogeneous signals as joint contributors to discover a shared semantic manifold.

\begin{takeawaybox}
\textbf{Positioning of Our Work:} UniM2 addresses the absence of labels by establishing \textbf{correspondence synergy} as an \textbf{intrinsic supervisory signal}. While previous methods rely on one-way imitation to import external guidance, UniM2 leverages a unified latent space as a mediator to enable \textbf{mutual supervision} between heterogeneous modalities, resolving structural contradictions and stabilizing the shared manifold without supervision.
\end{takeawaybox}

\label{sec:related}

\vspace{-10pt}
\section{Methodology}
\vspace{-10pt}
\subsection{Preliminaries and Task Definition}

\textbf{Preliminaries.} In the USS task, given an unlabeled image corpus $\mathcal{I} = \{I_i\}_{i=1}^N$, the objective is to learn a mapping function that assigns each pixel to one of the $K$ latent semantic clusters without any human supervision. State-of-the-art frameworks \cite{STEGO,EAGLE,HP} typically tackle this by distilling high-dimensional features from a frozen self-supervised backbone $\mathcal{F}$ \cite{DINOV3} into a compact, low-rank embedding space via a lightweight segmentation head $\mathcal{S}$ \cite{STEGO}. This semantic-preserving dimensionality reduction yields low-dimensional manifolds that are more amenable to clustering \cite{koenig2023uncovering}, as they mitigate the curse of dimensionality while amplifying latent semantic correspondences.

The core of this paradigm is a novel correspondence distillation loss \cite{xu2023self,fundel2025distillation, gou2021knowledge} that operates on three types of paired inputs: \textit{Self}, \textit{KNN} \cite{KNN1, knn2}, and \textit{Random} pairs \cite{STEGO}. For any given pair of images $(I_1, I_2)$, the framework extracts their dense feature maps $\{f_1, f_2\}$ from the frozen backbone $\mathcal{F}$, and generates the corresponding segmentation embeddings $\{s_1, s_2\}$ via the segmentation head $\mathcal{S}$ \cite{STEGO}. The underlying assumption is that the semantic correlation between $f_1$ and $f_2$ should be preserved and amplified in the embedding space of $s_1$ and $s_2$ \cite{STEGO}. Formally, the feature correspondence $F$ and segmentation correspondence $S$ are defined as pixel-wise cosine similarity \cite{cos1,cos2}:
\begin{equation}
F_{hwij} = \frac{f_{1,hw}^\top f_{2,ij}}{\|f_{1,hw}\| \|f_{2,ij}\|}, \quad S_{hwij} = \frac{s_{1,hw}^\top s_{2,ij}}{\|s_{1,hw}\| \|s_{2,ij}\|},
\label{cos}
\end{equation}
where $(h,w)$ and $(i,j)$ denote spatial indices. For \textit{Self} pairs, the image subscripts are omitted as the correspondence is computed within the same image ($I_1 = I_2$). The distillation process is then driven by the objective:
\begin{equation}
\mathcal{L} = -\sum_{hwij} (F_{hwij} - b) \odot \max(S_{hwij}, 0),
\end{equation}
where $b$ is a scalar bias providing negative pressure to force weakly correlated features toward orthogonality. Its value is specifically conditioned on the pair type (\textit{Self}, \textit{KNN}, or \textit{Random})  to prevent representation collapse and encourage the formation of compact clusters.

\textbf{Definition of UMSS.} Building upon the USS, we formally define the UMSS task. Unlike USS, which operates on a single image corpus, UMSS leverages a paired dataset $\mathcal{D} = \{(I_i, \{X_i^{(m)}\}_{m=1}^M)\}_{i=1}^N$, where each RGB image $I_i$ is aligned with a set of $M$ auxiliary modalities. The objective is to learn a joint representation space that captures consistent semantic categories across these heterogeneous inputs without any human annotations. Specifically, the framework aims to optimize a multi-modal mapping $\Phi(f_{rgb}, \{f_X^{(m)}\}_{m=1}^M) \to s$, where $f_{rgb}$ and $f_X^{(m)}$ denote dense features extracted from the RGB and the $m$-th auxiliary modality, respectively. However, as demonstrated by our preliminary experiments in Fig. \ref{fig:preliminary_study}, directly adopting existing multi-modal fusion schemes in this context proves counterproductive and even degrades performance. This failure highlights the inherent difficulty of aligning heterogeneous features without proper regularization. To address this, we propose \textbf{UniM2}, a framework designed to effectively harness multi-source information and resolve modality conflicts.

\vspace{-10pt}
\subsection{The Proposed UniM2 Framework}
\label{sec:UniM2}
\begin{figure*}[t]
  \centering
  \includegraphics[width=\linewidth]{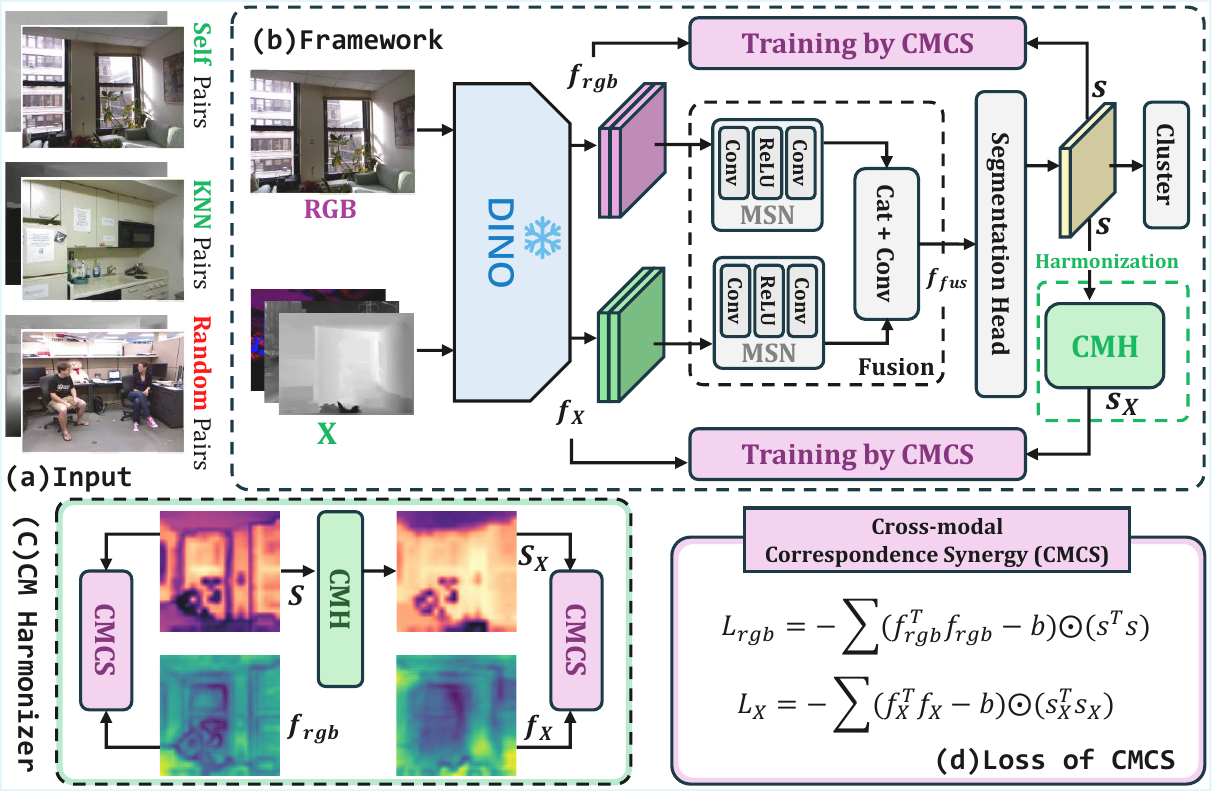}
  \vspace{-15pt}
\caption{\textbf{UniM2 Framework Overview.} (a) Training inputs for UniM2, (b) The overall architecture of UniM2, (c) The Cross-modal Harmonization process, and (d) the formulation of the CMCS loss. For illustration, (c) and (d) are depicted based on the \textit{Self} pair scenario to resolve cross-modal structural contradictions.}
\vspace{-16pt}
\label{fig:framework}
\end{figure*}

The overall architecture of UniM2 is illustrated in Fig. \ref{fig:framework} (b). \textbf{Taking the RGB-Depth pair as a representative case}, our framework differentiates itself from conventional USS by processing dual-modality inputs for each sample, denoted as $\{I, X\}$. As shown in the framework, both the RGB image $I$ and the auxiliary modality $X$ are first fed into a frozen self-supervised backbone $\mathcal{F}$ to extract their respective dense feature maps, $f_{rgb}$ and $f_X$. These features are subsequently processed by Modality-Specific Networks (MSN) for refinement before being integrated through a Conv Fusion module to generate the final fused representation $f_{fus}$. Finally, $f_{fus}$ is projected into a compact embedding space via $\mathcal{S}$ and subsequently clustered to yield the final semantic assignments.

\textbf{Training Inputs.} To optimize the framework, we adopt and extend the sampling strategy of \cite{STEGO} to a multi-modal context as shown in Fig. \ref{fig:framework} (a). In the original unimodal USS setting, a training pair consists of two images $(I_1, I_2)$, resulting in two backbone feature maps $\{f_1, f_2\}$ which are then mapped to two segmentation embeddings $\{s_1, s_2\}$ via a segmentation head $S$. In contrast, our UMSS approach operates on multi-modal groups, where a training pair comprises a source group $\mathcal{G}_1 = \{I_1, X_1\}$ and a target group $\mathcal{G}_2 = \{I_2, X_2\}$. Depending on the relationship between $\mathcal{G}_1$ and $\mathcal{G}_2$, the pair forms a \textit{Self}, \textit{KNN}, or \textit{Random} correspondence. Consequently, each training pair in UniM2 generates four distinct backbone feature maps, $\{f_{rgb,1}, f_{X,1}, f_{rgb,2}, f_{X,2}\}$, yet still results in only two final segmentation embeddings $\{s_1, s_2\}$. Here, each $s_i$ is derived from the multi-modal mapping $\Phi$, which is implemented as $s_i = \Phi(f_{rgb,i}, f_{X,i}) = S(\Psi(MSN(f_{rgb,i}), MSN(f_{X,i})))$. In this formulation, $\Psi$ represents the learnable fusion module that integrates heterogeneous features, while $S$ denotes the segmentation head that projects the fused representation into the low-dimensional manifolds. The core objective of this framework is to optimize and activate the learnable fusion module $\Psi$ without human supervision \textit{by enforcing structural consistency between the fused embeddings and the heterogeneous features from frozen backbones.}

\textbf{Modality Fusion.} We integrate the refined features via a \textbf{learnable Conv Fusion} module  $\Psi$: $f_{fus} = \Psi(MSN(f_{rgb}), MSN(f_{X}))$. While learnable fusion often converges to trivial solutions in unsupervised settings \cite{un_fusion1,un_fusion2}, we demonstrate that $\Psi$ can significantly outperform static operations (\eg, sum or average) when properly constrained. In UniM2, this integration is  regularized by the \textbf{Cross-modal Correspondence Synergy (CMCS)}, which provides the explicit guidance necessary to harness cross-modal synergies.

\textbf{Training by CMCS.} 
The core philosophy of our training objective extends the principle of correspondence distillation to the multi-modal domain. While conventional USS \cite{STEGO} assumes that semantic similarity in a low-dimensional space should mirror the correlations of a single modality, we instead aim to achieve a \textbf{cross-modal relational consensus}. We posit that a robust unified semantic space should preserve only those structural relationships that are consistently supported across its constituent modalities. Specifically, if a semantic relationship is jointly captured by heterogeneous sensors, the shared embedding space should reflect this agreement through consistent cross-modal correspondences. Accordingly, we propose \textbf{Cross-modal Correspondence Synergy}, which encourages the fused embeddings $s$ to be jointly constrained by the structural correlations of both $f_{rgb}$ and $f_X$. By emphasizing multi-source agreement rather than one-way imitation, our framework facilitates the discovery of a shared semantic manifold while suppressing modality-specific noise.

Specifically, for a training pair $(\mathcal{G}_1, \mathcal{G}_2)$, we extract the dense feature maps $f_{rgb,1}, f_{rgb,2}$ and $f_{X,1}, f_{X,2}$ directly from the frozen backbone to serve as stable semantic anchors. The modality-specific correspondence tensors, $F^{rgb}$ and $F^{X}$, are then defined as:
\begin{equation}
F^{rgb}_{hwij} = \cos(f_{rgb,1,hw}, f_{rgb,2,ij}), \quad F^{X}_{hwij} = \cos(f_{X,1,hw}, f_{X,2,ij}),
\end{equation}
where $(h,w)$ and $(i,j)$ denote the spatial coordinates. Note that these cosine similarities are computed following the same formulation as in Eq.~\ref{cos}. The objective of CMCS is to distill the structural correlations from individual modalities into the unified semantic space. Similarly, the correspondence in the unified semantic space is computed as:
\begin{equation}
    S_{hwij} = \cos(s_{1,hw}, s_{2,ij}),
\end{equation}
where $s_1$ and $s_2$ denote the final segmentation embeddings of the two groups. To align the unified semantic space with its constituent modalities, the total CMCS loss is formulated as the weighted sum of individual modality-specific losses:
\begin{equation}
    \mathcal{L}_{cmcs} = \mathcal{L}_{rgb} + \lambda \mathcal{L}_{X},
    \label{eq:6}
\end{equation}
where $\lambda$ is a balancing hyperparameter. For the RGB and auxiliary modalities, the specific loss terms $\mathcal{L}_{rgb}$ and $\mathcal{L}_{X}$ are formulated as:
\begin{equation}
    \resizebox{0.9\linewidth}{!}{$\displaystyle
        \mathcal{L}_{rgb} = - \sum_{h,w,i,j} (F^{rgb}_{hwij} - b) \odot \max(0, S_{hwij}), \quad \mathcal{L}_{X} = - \sum_{h,w,i,j} (F^{X}_{hwij} - b) \odot \max(0, S^{X}_{hwij}),
    $}
\label{eq:7}
\end{equation}
where $S^{X}_{hwij}$ represents the embedding correspondence processed by the \textbf{Cross-modal Harmonizer (CMH)}, which will be detailed in the following section.


\subsection{Cross-modal Harmonization for Modality Conflict}
\label{sec:CMH}

\begin{figure}[t]
    \centering
    \includegraphics[width=\linewidth]{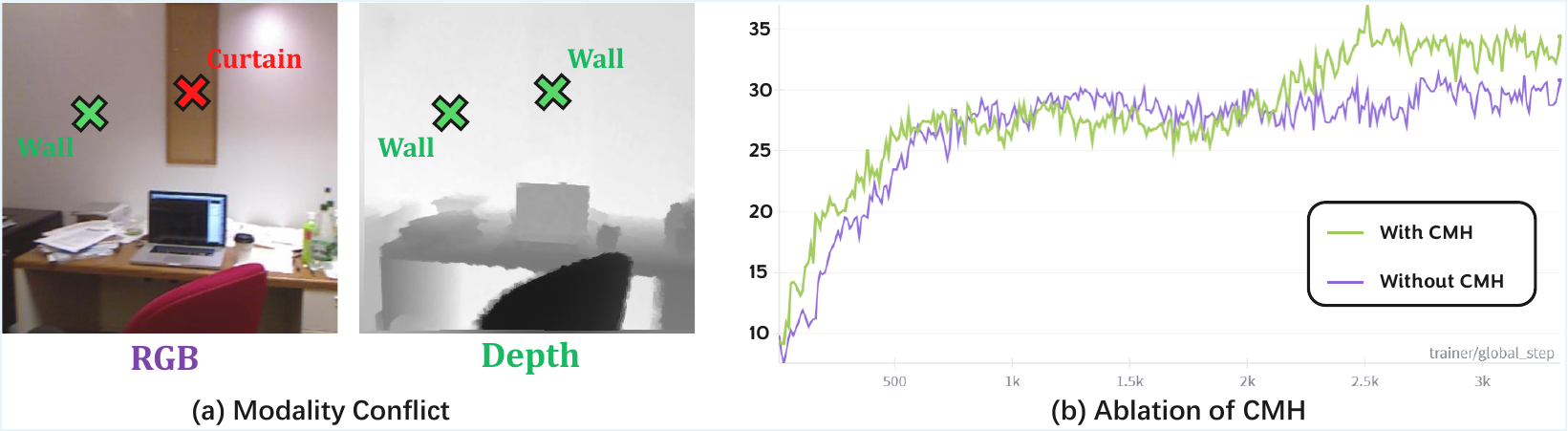}
    \caption{\textbf{Analysis of modality conflicts and CMH efficacy.} (a) Example of contradictory semantic cues between RGB (semantic) and depth (geometry). (b) mIoU performance comparison with and without the proposed CMH.}
    \label{fig:conflict}
    \vspace{-15pt}
\end{figure}

Despite its efficacy, CMCS relies on an implicit assumption of cross-modal semantic consistency, which is frequently violated by inherent physical sensor differences. As illustrated in Fig.~\ref{fig:conflict} (a), RGB sensors capture sharp textural boundaries for objects like curtains, whereas depth sensors may perceive them as part of the wall due to negligible depth variance. Such contradictions introduce conflicting gradients that confuse the optimization process and degrade the quality of the unified embedding space $s$. 

To mitigate these conflicts, we propose the \textbf{Cross-modal Harmonization} mechanism. Instead of enforcing a rigid, direct alignment between the unified embedding $s$ and auxiliary features $f_X$, we designate RGB as the primary semantic reference and decouple the auxiliary supervision via a learnable buffer:
\begin{equation}
    s^X = \text{CMH}(s),
\end{equation}
where $\text{CMH}(\cdot)$ is a lightweight learnable transformation. This buffer serves as a flexible mediation layer that facilitates a \textit{soft-alignment} between the unified space and auxiliary modalities. The harmonized correspondence $S^X_{hwij}$ used in Eq.~\ref{eq:7} is then formulated as:
\begin{equation}
    S^X_{hwij} = \cos(s^X_{1,hw}, s^X_{2,ij}).
\end{equation}
By supervising the transformed embedding $s^X$ rather than the shared space $s$ with the auxiliary signal $f_X$, CMH adaptively absorbs complementary cues while insulating the primary semantic manifold from pixel-wise modality contradictions. In practice, while CMH can be instantiated as any learnable network, we implement a lightweight two-layer convolutional structure for simplicity. This choice regulates the \textit{strictness of alignment}: the capacity of this mediation layer determines the degree to which the primary space $s$ is constrained by the auxiliary signal $f_X$. Such buffering ensures that auxiliary modalities provide structural guidance without distorting the unified semantic manifold. As evidenced in Fig.~\ref{fig:conflict} (b), this mechanism achieves remarkable mIoU gains.

\textbf{Scalability to More Modalities.} 
The modularity of CMH enables UniM2 to scale seamlessly to multiple auxiliary modalities. By assigning an independent CMH branch to each source, auxiliary supervision is decoupled from the shared space $s$, preventing gradient interference. This allows the unified embedding to be jointly regularized without complex balancing strategies. Accordingly, the total CMCS loss generalizes to:
\begin{equation}
    \mathcal{L}_{cmcs} = \mathcal{L}_{rgb} + \sum_{n=1}^{N} \lambda_n \mathcal{L}_{X_n},
\end{equation}
where $\mathcal{L}_{X_n}$ is the harmonized loss for the $n$-th modality. Mediating signals through independent transformations ensures stable optimization even as $N$ increases, allowing UniM2 to scale across diverse sensing configurations while preserving the integrity of the primary semantic manifold.

\label{sec:med}

\section{Experiments}

\label{sec:experiments}

\subsection{Experimental Setup}

\noindent \textbf{Datasets.} We evaluate UniM2 on three representative multi-modal benchmarks: NYU-Depth-V2 \cite{NYU_Depth}, MFNet \cite{MFNET}, and MCubeS \cite{McubeS}.

\noindent \textbf{NYU-Depth-V2} \cite{NYU_Depth} is a standard indoor RGB-D segmentation benchmark containing 1,449 aligned RGB–depth image pairs with dense annotations. We adopt the 13 class evaluation protocol for indoor semantic segmentation.

\noindent \textbf{MFNet} \cite{MFNET} is an urban RGB–thermal segmentation dataset with 1,569 aligned image pairs captured under both daytime and nighttime. We evaluate performance on its 8 categories to assess robustness under illumination changes.

\noindent \textbf{MCubeS} \cite{McubeS} is a quad-modal dataset for semantic material segmentation, featuring aligned RGB, Near-Infrared (NIR), Degree of Linear Polarization (DoLP), and Angle of Linear Polarization (AoLP) images. We perform evaluation on its 20 categories to validate our model's effectiveness in \textbf{fusing more than two modalities} for robust material recognition. 


\noindent \textbf{Implementation Details.}
We employ \textbf{DINOv3} as the frozen backbone. Following \cite{STEGO}, we adopt a \textit{5-crop} strategy during pre-processing to enhance spatial resolution and feature correspondence quality. The segmentation head consists of two convolutional layers with an intermediate activation layer, consistent with standard USS frameworks. All ablation studies are conducted using the \textbf{DINOv3-Small/16} variant. We use mean Intersection over Union (\textbf{mIoU}) and Pixel Accuracy (\textbf{Acc.}) as our evaluation metrics. Benefiting from the frozen backbone, the training process remains efficient, with a total training time of \textbf{less than two hours per model} on a single NVIDIA GeForce RTX 5090 GPU.

\noindent \textbf{Hyperparameter Settings and Fairness.}
Unsupervised semantic segmentation is generally sensitive to hyperparameter choices, and this sensitivity is not unique to UniM2. To ensure a fair and reproducible comparison, we allocate the same hyperparameter search budget to all compared methods. Specifically, each method is tuned with \textbf{200 iterations of Bayesian hyperparameter optimization} \cite{wu2019hyperparameter}, rather than using a single unified configuration across different datasets and methods, which can lead to suboptimal or biased results. All models are trained using the \textbf{Adam} optimizer with a learning rate of $5 \times 10^{-4}$ and a batch size of 32. Additional hyperparameter details, including those related to $\lambda$ and $b$, are provided in the \textbf{Supplementary Material}.

\subsection{Comparison Results}

\noindent \textbf{Comparison Methods.}
Our primary comparisons are based on USS extensions, including direct K-means clustering on DINO features, representative USS methods such as STEGO~\cite{STEGO} and EAGLE~\cite{EAGLE}, and their multi-modal variants. These baselines are the most relevant to UMSS, as they share the same label-free semantic segmentation objective and can be directly extended to multi-modal inputs. We also include image-level fusion and RGB-to-X distillation alternatives as supplementary comparisons. The former is constrained by modality-specific input compatibility, while the latter follows an asymmetric imitation protocol rather than joint multi-modal representation learning. The overall comparison coverage is summarized in Tab.~\ref{tab:comparison_summary}, with detailed supplementary results provided in the \textbf{Supplementary Material}.

\begin{table}[t]
\centering
\caption{\textbf{Summary of comparison coverage.} We compare UniM2 with representative alternatives from image-level fusion, RGB-to-X distillation, and USS-based multi-modal extensions. The reported values are representative mIoU results: NYU-Depth-V2 uses DINOv3-Small/16, while MFNet uses DINOv3-Base/16.}
\label{tab:comparison_summary}
\footnotesize
\setlength{\tabcolsep}{5.0pt}
\renewcommand{\arraystretch}{1.08}
\makebox[\linewidth][c]{
\scalebox{0.88}{
\begin{tabular}{llccc}
\toprule
\textbf{Category} & \textbf{Method} & \textbf{NYU-Depth-V2} & \textbf{MFNet} & \textbf{Location} \tabularnewline
\midrule
Image fusion & SwinFusion~\cite{ma2022swinfusion} & - & 36.2 & Supp. Chapter 1 \tabularnewline
Image fusion & Mask-DiFuser~\cite{tang2025mask} & - & 39.1 & Supp. Chapter 1 \tabularnewline
\midrule
RGB to X distillation & CORAL~\cite{coral} & 21.3 & - & Supp. Chapter 2 \tabularnewline
RGB to X distillation & MMD~\cite{mmd} & 20.1 & - & Supp. Chapter 2 \tabularnewline
RGB to X distillation & Cosine & 24.5 & - & Supp. Chapter 2 \tabularnewline
\midrule
USS extension & STEGO~\cite{STEGO} & 28.8 & 35.9 & Main Text \tabularnewline
USS extension & EAGLE~\cite{EAGLE} & 27.4 & 37.8 & Main Text \tabularnewline
\midrule
\textbf{Ours} & \textbf{UniM2} & \textbf{36.9} & \textbf{45.7} & Main Text \tabularnewline
\bottomrule
\end{tabular}
}
}
\vspace{-2mm}
\end{table}

\noindent \textbf{Performance on NYU-Depth-v2 and MFNet.}
Tab.~\ref{tab:nyu_mfnet_results} shows that directly introducing auxiliary modalities into existing USS baselines does not necessarily improve performance. For both STEGO and EAGLE, naive Depth/Thermal fusion often leads to clear mIoU degradation, suggesting that heterogeneous modalities introduce structural conflicts that cannot be properly resolved without label supervision. In contrast, UniM2 consistently turns auxiliary modalities into positive gains. With DINOv3-Base/16, UniM2 improves over the RGB-only STEGO baseline by 6.4 mIoU on NYU-Depth-v2 and 9.8 mIoU on MFNet. These results demonstrate that CMCS provides effective cross-modal correspondence supervision, while CMH mitigates unreliable auxiliary guidance and stabilizes multi-modal representation learning.

\noindent \textbf{Per-class Analysis on NYU-Depth-v2.}
Tab.~\ref{tab:per_class_results_wide} provides a finer-grained view of how different fusion strategies affect semantic categories. Naive depth fusion improves geometry-sensitive categories such as \textit{Sofa}, where depth offers useful structural cues, but it substantially hurts appearance-dominant categories such as \textit{Floor} and \textit{Wall}. This indicates that auxiliary modalities can be beneficial for some categories while being harmful for others if cross-modal conflicts are not controlled. UniM2 achieves a better balance: it obtains the best results on \textit{Sofa}, \textit{Table}, and \textit{TV}, while preserving strong performance on \textit{Floor} and \textit{Wall}. This confirms that UniM2 can exploit complementary geometric information without sacrificing the semantic structure captured by RGB.

\begin{table}[t]
\centering
\caption{\textbf{Quantitative comparison on NYU-Depth-v2 \cite{NYU_Depth} and MFNET \cite{MFNET} datasets.} Note that performance variations ($\uparrow$, $\downarrow$) for UniM2 are reported relative to the RGB-only baseline of STEGO \cite{STEGO}.}
\vspace{-10pt}
\label{tab:nyu_mfnet_results}
\setlength{\tabcolsep}{3pt} 
\renewcommand{\arraystretch}{1.1} 

\resizebox{\linewidth}{!}{
\begin{tabular}{l c c r l r l r l r l}
\toprule
\multirow{2}{*}{Method} & \multirow{2}{*}{Modality} & \multirow{2}{*}{Backbone} & \multicolumn{4}{c}{NYU-Depth-v2 \cite{NYU_Depth}} & \multicolumn{4}{c}{MFNET \cite{MFNET}} \\
\cmidrule(lr){4-7} \cmidrule(lr){8-11}
& & & \multicolumn{2}{c}{mIoU $\uparrow$} & \multicolumn{2}{c}{Acc. $\uparrow$} & \multicolumn{2}{c}{mIoU $\uparrow$} & \multicolumn{2}{c}{Acc. $\uparrow$} \\
\midrule
\multirow{2}{*}{DINOv3 \cite{DINOV3}} & RGB & \multirow{7}{*}{ViT-S/16} & 11.1 & & 26.2 & & 20.1 & & 67.3 & \\
& + Depth/Thermal & & 9.4 & \textcolor{red}{\scriptsize ($\downarrow$ 1.7)} & 25.5 & \textcolor{red}{\scriptsize ($\downarrow$ 0.7)} & 19.8 & \textcolor{red}{\scriptsize ($\downarrow$ 0.3)} & 66.5 & \textcolor{red}{\scriptsize ($\downarrow$ 0.8)} \\ \cmidrule(lr){1-2}

\multirow{2}{*}{+ STEGO \cite{STEGO}} & RGB & & 28.8 & & 52.0 & & 32.2 & & 72.1 & \\
& + Depth/Thermal & & 25.3 & \textcolor{red}{\scriptsize ($\downarrow$ 3.5)} & 45.9 & \textcolor{red}{\scriptsize ($\downarrow$ 6.1)} & 31.3 & \textcolor{red}{\scriptsize ($\downarrow$ 0.9)} & 74.9 & \textcolor{ForestGreen}{\scriptsize ($\uparrow$ 2.8)} \\ \cmidrule(lr){1-2}

\multirow{2}{*}{+ EAGLE \cite{EAGLE}} & RGB & & 27.4 & & 51.6 & & 34.7 & & 79.6 & \\
& + Depth/Thermal & & 20.1 & \textcolor{red}{\scriptsize ($\downarrow$ 7.3)} & 40.0 & \textcolor{red}{\scriptsize ($\downarrow$ 11.6)} & 31.1 & \textcolor{red}{\scriptsize ($\downarrow$ 3.6)} & 77.4 & \textcolor{red}{\scriptsize ($\downarrow$ 2.2)} \\ \cmidrule(lr){1-2}

\rowcolor[gray]{0.95} 
\textbf{+ UniM2 (Ours)} & \textbf{+ Depth/Thermal} & & \textbf{36.9} & \textcolor{ForestGreen}{\scriptsize ($\uparrow$ 8.1)} & \textbf{56.1} & \textcolor{ForestGreen}{\scriptsize ($\uparrow$ 4.1)} & \textbf{35.2} & \textcolor{ForestGreen}{\scriptsize ($\uparrow$ 3.0)} & \textbf{81.5} & \textcolor{ForestGreen}{\scriptsize ($\uparrow$ 9.4)} \\

\midrule 

\multirow{2}{*}{DINOv3 \cite{DINOV3}} & RGB & \multirow{7}{*}{ViT-B/16} & 14.3 & & 32.8 & & 20.0 & & 72.1 & \\
& + Depth/Thermal & & 10.4 & \textcolor{red}{\scriptsize ($\downarrow$ 3.9)} & 23.3 & \textcolor{red}{\scriptsize ($\downarrow$ 9.5)} & 21.2 & \textcolor{ForestGreen}{\scriptsize ($\uparrow$ 1.2)} & 72.4 & \textcolor{ForestGreen}{\scriptsize ($\uparrow$ 0.3)}\\ \cmidrule(lr){1-2}

\multirow{2}{*}{+ STEGO \cite{STEGO}} & RGB & & 31.7 & & 55.1 & & 35.9 & & 73.6 & \\
& + Depth/Thermal & & 31.1 & \textcolor{red}{\scriptsize ($\downarrow$ 0.6)}  & 49.7 & \textcolor{red}{\scriptsize ($\downarrow$ 5.4)} & 32.5 & \textcolor{red}{\scriptsize ($\downarrow$ 3.4)} & 74.1 & \textcolor{ForestGreen}{\scriptsize ($\uparrow$ 0.5)}\\ \cmidrule(lr){1-2}

\multirow{2}{*}{+ EAGLE \cite{EAGLE}} & RGB & & 30.9 & & 49.5 & & 37.8& & 72.5& \\
& + Depth/Thermal & & 25.8 & \textcolor{red}{\scriptsize ($\downarrow$ 5.1)} & 46.9 & \textcolor{red}{\scriptsize ($\downarrow$ 2.6)} & 33.5 & \textcolor{red}{\scriptsize ($\downarrow$ 4.3)} & 74.5 & \textcolor{ForestGreen}{\scriptsize ($\uparrow$ 2.0)} \\ \cmidrule(lr){1-2}

\rowcolor[gray]{0.95} 
\textbf{+ UniM2 (Ours)} & \textbf{+ Depth/Thermal} & & \textbf{38.1} & \textcolor{ForestGreen}{\scriptsize ($\uparrow$ 6.4)} & \textbf{58.8} & \textcolor{ForestGreen}{\scriptsize ($\uparrow$ 3.7)} & \textbf{45.7} & \textcolor{ForestGreen}{\scriptsize ($\uparrow$ 9.8)} & \textbf{76.1} & \textcolor{ForestGreen}{\scriptsize ($\uparrow$ 3.7)} \\ 

\bottomrule
\end{tabular}
}

\end{table}

\noindent \textbf{Performance on MCubeS.}
Tab.~\ref{tab:mcubes_results} further evaluates UniM2 under a more challenging multi-modal setting with RGB, NIR, DoLP, and AoLP inputs. UniM2 achieves consistent improvements when informative modalities are added, such as $I \to IN$ and $ID \to IND$, demonstrating that the proposed CMH design can naturally extend beyond bi-modal fusion. Meanwhile, the slight drops observed in settings involving AoLP suggest that weak or noisy modalities may still limit the final performance. This observation is consistent with supervised multi-modal segmentation, where sensor quality and modality reliability remain important factors~\cite{liao2025benchmarking, MAGIC, zheng2025reducing}.

\begin{figure}[t]
\centering
\includegraphics[width=\linewidth]{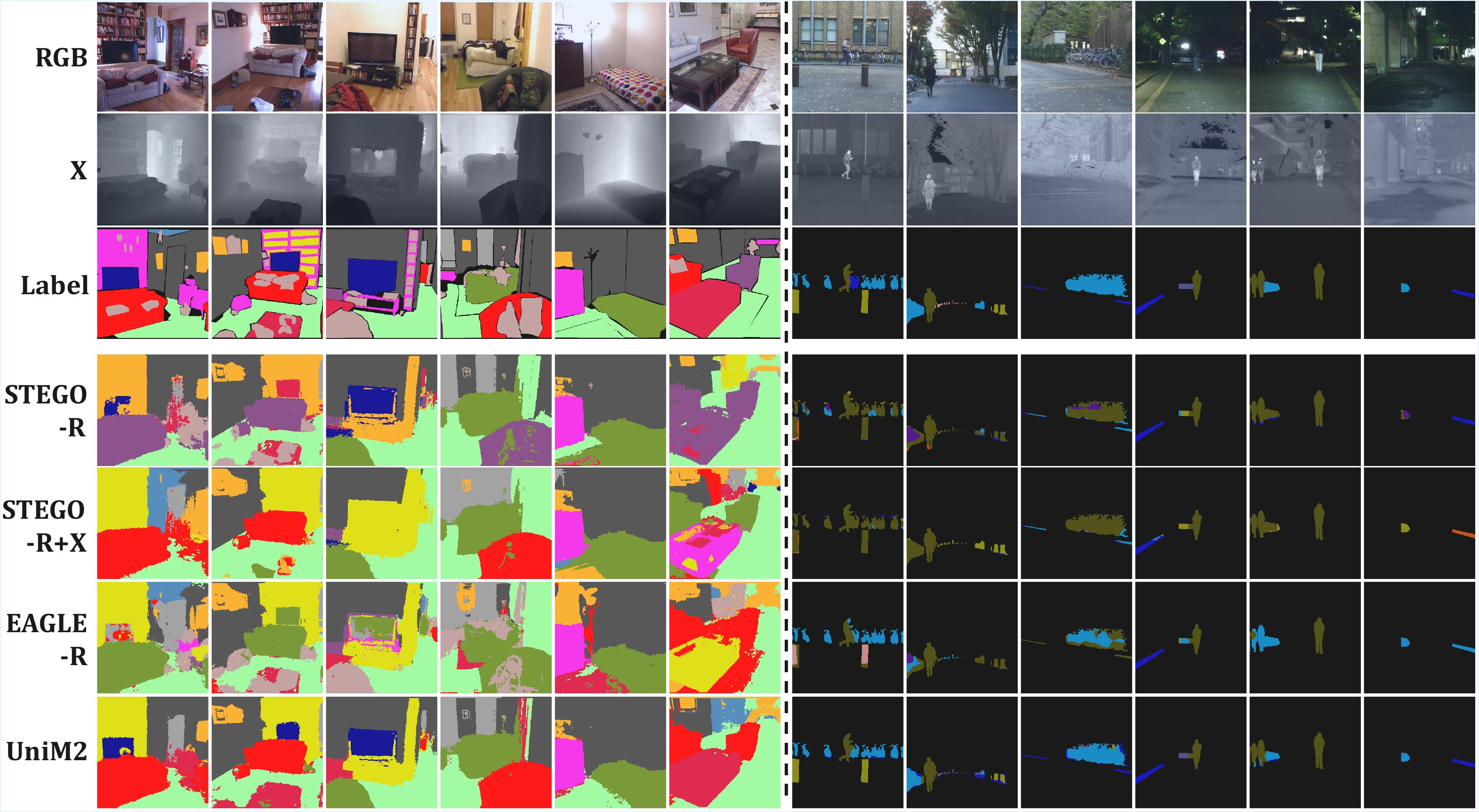}
\caption{\textbf{Qualitative Comparison.} Visual results on the NYU-Depth-v2 and MFNet datasets, comparing UniM2 against RGB-only and multi-modal (R+X) variants of STEGO~\cite{STEGO} and RGB-only EAGLE~\cite{EAGLE} baselines.}
\label{fig:qualitative_comparison}
\vspace{-5mm}
\end{figure}

\begin{figure}[t]
\centering
\includegraphics[width=\linewidth]{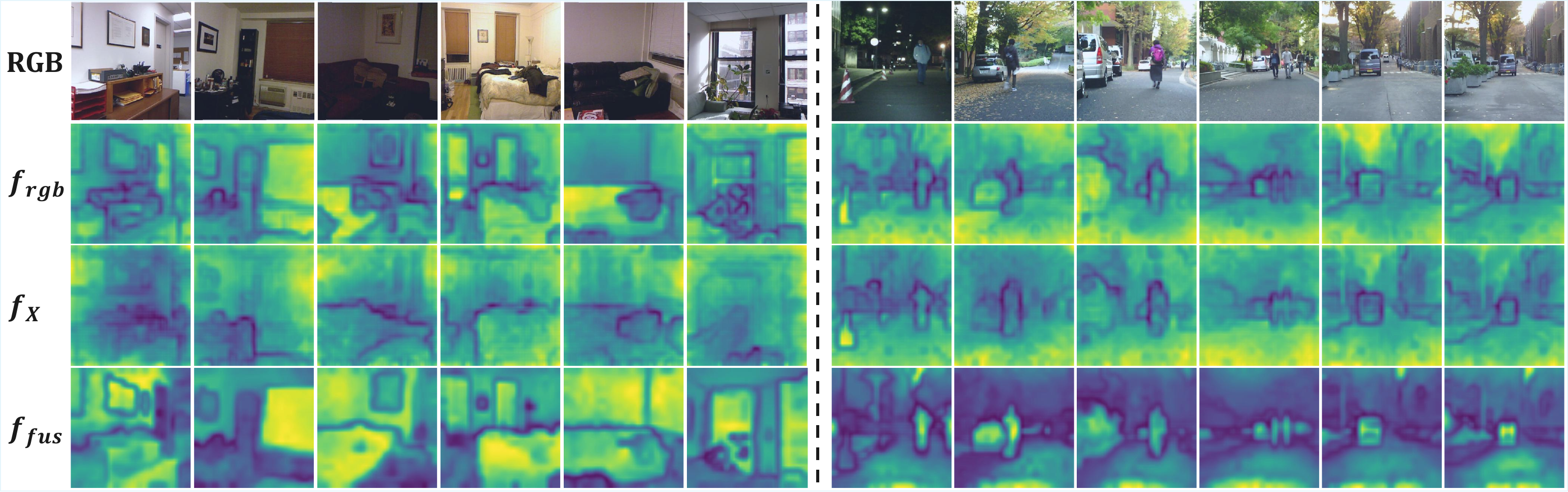}
\caption{\textbf{Feature Visualization.} Comparative visualization of feature maps from the RGB modality, the auxiliary modality, and the proposed fused representation.}
\label{fig:vis_features}
\vspace{-2mm}
\end{figure}

\noindent \textbf{Visualization Results.}
Fig.~\ref{fig:qualitative_comparison} visually compares UniM2 with RGB-only and naive multi-modal baselines on NYU-Depth-v2 and MFNet. While baseline methods often produce fragmented masks or incorrect regions after introducing auxiliary modalities, UniM2 generates cleaner semantic maps with sharper object boundaries. Fig.~\ref{fig:vis_features} further shows that the fused representation $f_{fus}$ is more spatially coherent than individual modality features, qualitatively supporting the effectiveness of CMCS and CMH in resolving modality conflicts.

\begin{table*}[t]
\centering
\caption{Per-class IoU comparison on the NYU-Depth-v2 dataset. The \textbf{best} and \underline{second-best} results are highlighted in \textbf{bold} and \underline{underline}, respectively.}
\vspace{-10pt}
\label{tab:per_class_results_wide}
\setlength{\tabcolsep}{3.5pt} 
\renewcommand{\arraystretch}{1.1} 

\resizebox{\textwidth}{!}{
\begin{tabular}{l c | c c c c c c c c c c c c c | c}
\toprule
Method & Modality & Bed & Book & Ceil & Chair & Floor & Furn & Obj & Pic & Sofa & Table & TV & Wall & Wind & \textbf{mIoU} \\
\midrule

\multirow{2}{*}{STEGO \cite{STEGO}} & RGB & 
\textbf{54.2} & 1.1 & \underline{25.1} & 34.6 & \underline{58.4} & \underline{44.7} & 19.2 & 19.8 & 1.0 & \underline{15.8} & \underline{13.6} & \textbf{55.8} & \textbf{68.8} & \underline{31.7} \\
& + Depth & 
51.2 & 11.4 & 21.8 & 41.4 & 23.3 & 41.5 & 15.5 & 29.8 & \underline{46.7} & 14.5 & 9.4 & 40.9 & 56.9 & 31.1 \\
\midrule

\multirow{2}{*}{EAGLE \cite{EAGLE}} & RGB & 
44.9 & \underline{15.5} & \textbf{46.0} & \underline{41.5} & 45.7 & 43.8 & \textbf{23.3} & \textbf{31.9} & 0.0 & 9.1 & 0.0 & 43.0 & 57.0 & 30.9 \\
& + Depth& 
47.3 & \textbf{16.3} & 2.1 & 30.4 & \textbf{68.4} & \textbf{45.5} & 19.2 & 6.4 & 1.6 & 4.7 & 5.8 & 39.6 & 48.1 & 25.8 \\
\midrule

\rowcolor[gray]{0.95} 
\textbf{UniM2 (Ours)} & \textbf{+ Depth} & 
\underline{52.8} & 12.9 & 22.8 & \textbf{41.7} & 58.2 & 44.5 & \underline{21.8} & \underline{30.8} & \textbf{55.7} & \textbf{18.9} & \textbf{20.7} & \underline{53.5} & \underline{61.0} & \textbf{38.1} \\
\bottomrule
\end{tabular}
}
\vspace{-4pt}
\end{table*}

\subsection{Ablation Studies}
We conduct ablation studies on NYU-Depth-v2 with the \textbf{DINOv3-Small/16} model, covering component effectiveness, CMH placement, and fusion strategies.

\textbf{Component Effectiveness.} As shown in Tab.~\ref{tab:ablation_study}, the baseline without the proposed modules obtains 25.0\% mIoU, corresponding to multi-modal STEGO~\cite{STEGO} with Conv Fusion. Adding CMCS and MSN improves the result to 31.3\%, already surpassing the RGB-only baseline of 28.8\%. Introducing CMH further boosts the performance from 31.3\% to 36.9\%, highlighting its importance in harmonizing modality conflicts. The gain from 34.2\% to 36.9\% also confirms the benefit of MSN for feature refinement before fusion.

\begin{table}[t]
\centering
\caption{Quantitative comparison on the MCubeS \cite{McubeS} dataset. All methods utilize the ViT-S/16 backbone. I, N, A, and D denote RGB, NIR, AoLP, and DoLP modalities.}
\vspace{-10pt}
\label{tab:mcubes_results}
\setlength{\tabcolsep}{3.5pt} 
\renewcommand{\arraystretch}{1.1} 

\resizebox{\linewidth}{!}{
\begin{tabular}{l c c c c c c c c c c c c}
\toprule
\multirow{2}{*}{Method} & \multicolumn{2}{c}{I} & \multicolumn{2}{c}{IN} & \multicolumn{2}{c}{IA} & \multicolumn{2}{c}{ID} & \multicolumn{2}{c}{IND} & \multicolumn{2}{c}{INAD} \\
\cmidrule(lr){2-3} \cmidrule(lr){4-5} \cmidrule(lr){6-7} \cmidrule(lr){8-9} \cmidrule(lr){10-11} \cmidrule(lr){12-13}
& mIoU  & Acc. & mIoU & Acc.& mIoU & Acc. & mIoU & Acc. & mIoU & Acc. & mIoU & Acc. \\
\midrule
DINOv3 \cite{DINOV3} 
& 13.9 & 50.7 & 14.6 & 52.7 & 11.2 & 35.5 & 12.9 & 38.3 & 10.9 & 43.3 & 11.1 & 36.5 \\

+ STEGO \cite{STEGO} 
& 19.1 & 55.2 & 18.5 & 54.3 & 17.4 & 54.6 & 17.6 & 52.1 & 18.8 & 56.7 & 18.3 & 56.4 \\

+ EAGLE \cite{EAGLE} 
& 18.1 & 57.7 & 18.7 & 58.2 & 16.5 & 57.2 & 17.0 & 51.1 & 18.6 & 57.2 & 16.5 & 53.2 \\

\rowcolor[gray]{0.95} 
\textbf{+ UniM2 (Ours)} 
& - & - & \textbf{21.5} & \textbf{59.1} & \textbf{18.5} & \textbf{61.7} & \textbf{20.9} & \textbf{57.8} & \textbf{21.8} & \textbf{64.5} & \textbf{20.7} & \textbf{62.6} \\ 
\bottomrule
\end{tabular}
}

\end{table}



\textbf{Placement of CMH.} CMH provides learnable flexibility for mitigating structural conflicts, while the modality without CMH serves as a fixed anchor. Tab.~\ref{tab:cmh_position} studies this anchoring effect by applying CMH to different branches. The \textit{No anchors} setting drops to 19.8\% mIoU, showing that excessive flexibility makes training unstable without a reliable reference. Keeping \textit{RGB only} as the anchor achieves the best result of 36.9\%, clearly outperforming the \textit{Depth only} anchor setting of 27.6\%. This suggests that RGB offers a more reliable semantic structure for unsupervised clustering, while depth is better used as complementary guidance. This supports our asymmetric design, where RGB provides a stable semantic reference and the auxiliary modality adapts through CMH.

\textbf{Fusion Strategy.} Tab.~\ref{tab:fusion_ablation} compares different fusion operators. Conv Fusion achieves the highest mIoU of 36.9\%, outperforming static operations such as Max, Mean, and Sum. Unlike static fusion, which applies fixed aggregation rules, Conv Fusion can adaptively select useful cues across pixels and channels. These results show that learnable fusion can be effective in UMSS when modality conflicts are properly regulated by CMH.

\begin{takeawaybox}
\noindent More analyses are provided in the \textbf{Supplementary Material}, including: 
(1) \textbf{additional fusion baselines} \cite{ma2022swinfusion, tang2025mask, tang2025controlfusion} in UMSS; 
(2) in-depth \textbf{theoretical analysis of CMH}; 
(3) distillation paradigms used in CMKD/UDA for UMSS; 
(4) hyperparameter analysis in UMSS; and 
(5) per-category distribution and confusion matrices. 
Extensive visualizations are also provided.
\end{takeawaybox}

\vspace{-2mm}

\begin{table*}[t]
\centering
\scriptsize 
\renewcommand{\arraystretch}{1.15} 

\begin{minipage}[t]{0.38\linewidth}
    \centering
    \caption{Ablation of each component in UniM2.}
    \vspace{-10pt}
    \label{tab:ablation_study}
    \setlength{\tabcolsep}{2.5pt} 
    \begin{tabular}{c c c c c}
    \toprule
    CMCS & MSN & CMH & mIoU $\uparrow$ & Acc. $\uparrow$ \\
    \midrule
    &  & & 25.0 & 48.1 \\
    \checkmark & \checkmark & & 31.3 & 50.7 \\
    \checkmark & & \checkmark & 34.2 & 54.3 \\
    \rowcolor[gray]{0.95} \checkmark & \checkmark & \checkmark & \textbf{36.9} & \textbf{56.1} \\
    \bottomrule
    \end{tabular}
\end{minipage}
\hfill
\begin{minipage}[t]{0.30\linewidth}
    \centering
    \caption{Ablation of anchor position.}
    \vspace{-10pt}
    \label{tab:cmh_position}
    \setlength{\tabcolsep}{3pt}
    \begin{tabular}{l c c}
    \toprule
    Position & mIoU $\uparrow$ & Acc. $\uparrow$ \\
    \midrule
    Both & 31.3 & 50.7 \\
    Depth only & 27.6 & 45.3 \\
    \rowcolor[gray]{0.95} RGB only & \textbf{36.9} & \textbf{56.1} \\
    No anchors & 19.8 & 50.8 \\
    \bottomrule
    \end{tabular}
\end{minipage}
\hfill
\begin{minipage}[t]{0.28\linewidth}
    \centering
    \caption{Ablation of fusion strategies.}
    \vspace{-10pt}
    \label{tab:fusion_ablation}
    \setlength{\tabcolsep}{4pt}
    \begin{tabular}{l c c}
    \toprule
    Strategy & mIoU $\uparrow$ & Acc. $\uparrow$ \\
    \midrule
    Max & 30.5 & 50.2 \\
    Mean & 32.5 & 53.2 \\
    Sum  & 33.0 & 53.6 \\
    \rowcolor[gray]{0.95} Conv & \textbf{36.9} & \textbf{56.1} \\
    \bottomrule
    \end{tabular}
\end{minipage}
\vspace{-4mm}

\end{table*}

\label{sec:exp}

\section{Conclusion and Future Work}
In this paper, we have defined the task of \textbf{Unsupervised Multi-modal Semantic Segmentation} and proposed \textbf{UniM2}, a novel framework for leveraging heterogeneous modalities without any human annotations. We achieve effective multi-modal integration through \textbf{Cross-modal Correspondence Synergy}, which enforces structural consistency between a unified latent space and its constituent modalities. To address the inherent inter-modal conflicts arising from the diverse physical properties of different sensors, we further introduce the \textbf{Cross-modal Harmonizer}. By designating RGB as a stable semantic reference, CMH facilitates the absorption of complementary cues while mitigating contradictory supervision. \textbf{Crucially, UniM2 transforms unsupervised multi-modal semantic segmentation  from performance degradation to consistent and substantial gains}, reversing the common failure of conventional fusion schemes in label-free settings. Notably, our modular design ensures high \textbf{Scalability}, allowing UniM2 to effectively extend to multiple auxiliary modalities. We hope that our UniM2 framework and the established evaluation benchmark can serve as a valuable baseline and inspire further advancements in the field of UMSS. 

\clearpage  

\section*{Acknowledgements}

This work was supported by the MOE AcRF Tier 1 (Call 2/2025) Grant under Grant No. RG160/25, NTU Start-Up Grant, and NTU EEE Internal Funding.

%
%
\bibliographystyle{splncs04}
\bibliography{main}
\end{document}